\documentclass[11pt,a4paper]{article}
\usepackage[hyperref]{emnlp2020}
\usepackage{times}
\usepackage{latexsym}

\usepackage{xspace}
\usepackage{tabularx}

\newcommand{\ttbf}[1]{\texttt{\textbf{#1}}}
\newcommand{\nocaps}[0]{\emph{nocaps}\xspace}
\newcommand{\nbtpp}[0]{\ttbf{ECOL-R}\xspace}
\newcommand{\nbtppnor}[0]{\ttbf{ECOL}\xspace}
\newcommand{\coco}[0]{COCO\xspace}

\newcommand{\copyable}[0]{{task specific}\xspace}
\newcommand{\noncopyable}[0]{{Visual Genome}\xspace}

\newcommand{\openimages}[0]{\ttbf{Open Images}\xspace}
\newcommand{\updown}[0]{\ttbf{Up-Down}\xspace}
\newcommand{\nbt}[0]{\ttbf{NBT}\xspace}
\newcolumntype{C}[1]{>{\centering\arraybackslash}p{#1}}
\newcommand{\nocapsin}[0]{\emph{in-domain}\xspace}
\newcommand{\nocapsnear}[0]{\emph{near-domain}\xspace}
\newcommand{\nocapsout}[0]{\emph{out-of-domain}\xspace}

\newcolumntype{Y}{>{\centering\arraybackslash}X}
\newcolumntype{L}{>{\raggedright\arraybackslash}X}

\newcommand{\expectwrt}[2][]{\ensuremath{\textbf{E}_{#1}\!\left[#2\right]}}
\newcommand{\var}[1]{\mathit{#1}}
\mathchardef\mhyphen="2D
\usepackage{amsmath,bm}
\usepackage{amsfonts}

\def\eqref#1{equation~\ref{#1}}
\def\1{\bm{1}}

\def\vc{{\bm{c}}}

\def\ve{{\bm{e}}}

\def\vh{{\bm{h}}}

\def\vp{{\bm{p}}}

\def\vr{{\bm{r}}}

\def\vv{{\bm{v}}}
\def\vw{{\bm{w}}}
\def\vx{{\bm{x}}}

\def\mW{{\bm{W}}}

\DeclareMathAlphabet{\mathsfit}{\encodingdefault}{\sfdefault}{m}{sl}
\SetMathAlphabet{\mathsfit}{bold}{\encodingdefault}{\sfdefault}{bx}{n}

\newcommand{\R}{\mathbb{R}}

\newcommand{\softmax}{\mathrm{softmax}}

\usepackage{graphicx}
\graphicspath{ {./image/} }
\usepackage{amsmath}
\usepackage{paralist,tabularx}
\usepackage{booktabs}
\usepackage{subfiles} 
\usepackage{microtype}
\usepackage[toc,page]{appendix}

\aclfinalcopy % Uncomment this line for the final submission
\title{\nbtpp: Encouraging Copying in Novel Object Captioning with Reinforcement Learning}

\author{Yufei Wang$^{1}$ \and Ian D. Wood$^{1}$ \and Stephen Wan$^{2}$ \and Mark Johnson$^{3}$ \\
Macquarie University, Sydney, Australia$^1$ \\
CSIRO Data61, Sydney, Australia$^{2}$ \\
Oracle Digital Assistant, Oracle Corporation$^{3}$ \\
\texttt{yufei.wang@students.mq.edu.au, ian.wood@mq.edu.au} \\
\texttt{stephen.wan@data61.csiro.au} \\
\texttt{mark.mj.johnson@oracle.com} \\
}

\date{}

\begin{document}
\maketitle
\begin{abstract}
Novel Object Captioning is a zero-shot Image Captioning task requiring describing objects not seen in the training captions, but for which information is available from external object detectors. The key challenge is to select and describe all salient detected novel objects in the input images. In this paper, we focus on this challenge and propose the \nbtpp model (\textbf{E}ncouraging \textbf{C}opying of \textbf{O}bject \textbf{L}abels with \textbf{R}einforced Learning), a copy-augmented transformer model that is encouraged to accurately describe the novel object labels. This is achieved via a specialised reward function in the SCST reinforcement learning framework~\cite{Rennie_2017_CVPR} that encourages novel object mentions while maintaining the caption quality. We further restrict the SCST training to the images where detected objects are mentioned in reference captions to train the \nbtpp model. We additionally improve our copy mechanism via \emph{Abstract Labels}, which transfer knowledge from known to novel object types, and a \emph{Morphological Selector}, which determines the appropriate inflected forms of novel object labels.  The resulting model sets new state-of-the-art on the \nocaps~\cite{nocaps2019} and \emph{held-out} \coco~\cite{Hendricks_2016_CVPR} benchmarks.

\end{abstract}

\section{Introduction}
\label{intro}
\begin{figure}[t!]
\centering
\includegraphics[width=0.48\textwidth]{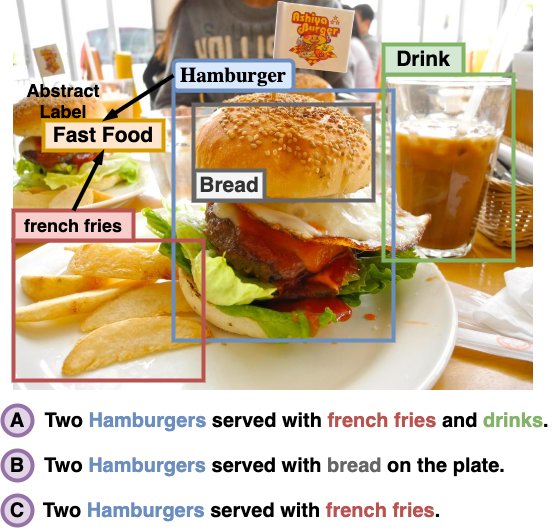}
\caption{Caption \textbf{A} is the ground-truth caption for the image. Compared with \textbf{B} and \textbf{C}, \textbf{A} is the best caption because it mentions all salient objects (i.e, \emph{Hamburger}, \emph{French Fries} and \emph{Drinks}). We use Abstract Labels, that is hypernyms of the objects' detected object labels in the object representations, transferring knowledge from the objects seen in the training captions to novel objects. Our copy mechanism also selects appropriate inflected forms of object labels (i.e., \emph{Hamburgers} vs. \emph{Hamburger}).}
\label{example}
\end{figure}

Novel Object Captioning is a zero-shot Image Captioning task where the captions should mention novel objects (i.e., not seen in the training captions), but for which information is available from external object detectors. To produce high-quality captions, the captioning models should select and describe all salient detected objects and avoid mentioning minor or irrelevant details in the input images. As shown in Figure~\ref{example}, caption \textbf{A} is the best caption among the three because \textbf{A} mentions all salient objects in the images without any unnecessary details while \textbf{B} mentions \emph{Bread} which is just a minor detail; and \textbf{C} misses the salient object \emph{Drink}. This paper aims to develop a captioning model that produces caption \textbf{A}.

We use an advanced copy mechanism, similar to the one in~\newcite{Hu_2020_CVPR}, to effectively integrate novel objects. We follow the setup in~\newcite{nocaps2019} and use two object detectors: one providing rich object visual features and another providing task specific (including novel) object labels as copy candidates. Our preliminary experiments show that the copy mechanism is infrequently triggered and unable to mention many salient objects in the input images. We propose the \nbtpp model (\textbf{E}ncouraging \textbf{C}opying of \textbf{O}bject \textbf{L}abels with \textbf{R}einforced Learning), a copy-augmented transformer model trained in the Self-Critical Sequence Training (SCST) framework~\cite{Rennie_2017_CVPR}. SCST with a CIDEr reward~\cite{Vedantam_2015_CVPR} is a standard approach for training the captioning models~\cite{Anderson_2018_CVPR}, but this paper will show that it does not sufficiently encourage the model to use copy operations. We design a new reward function that provides a reward for each copy operation proportional to the caption quality. We further restrict the SCST training to the images that contain at least one word in the ground truth captions that corresponds to one of the detected object labels. With these innovations, the \nbtpp model outperforms a SCST baseline and a strong inference encouragement baseline by a large margin.

Our copy mechanism and caption generator incorporate two enhancements to better choose and incorporate novel objects: \emph{a) Abstract Labels} which correspond to hypernyms of the object labels and facilitate knowledge transfer between objects appearing in training captions and novel objects; \emph{b) a Morphological Selector} which determines the correct inflected form of the copied \copyable object labels which is similar in purpose to that proposed in~\cite{lu-etal-2018-deep}. 

We evaluate the \nbtpp model on the novel object captioning benchmark \nocaps~\cite{nocaps2019} and \emph{held-out} \coco~\cite{Hendricks_2016_CVPR}. The \nbtpp model achieves a new state of the art on both benchmarks and generalizes well to in-domain images. 

\section{Related Work}

Popular Image Captioning models include LSTM-based~\cite{Anderson_2018_CVPR} and Transformer-based decoders~\cite{NIPS2019_9293,cornia2020m2}. The visual encoders are often neural object detectors~\cite{Anderson_2018_CVPR,wang2019hierarchical} producing Region-of-Interest (ROI) vectors. To train the model to copy novel object labels, the Neural Baby Talk model (\nbt)~\cite{lu2018neural} and follow-up work~\cite{wu2018decoupled,yao2017incorporating,Li_2019_CVPR} use copy mechanisms~\cite{NIPS2015_5866}. The copying candidates are labels of salient objects produced by external object detectors. In this paper, we follow previous work by using the \noncopyable object detector from~\cite{Anderson_2018_CVPR} as the visual feature extractor and a \copyable object detector to provide object labels for copying.

These models are typically trained with the Cross-Entropy loss (CE). This creates a mismatch between the training and testing environments because the evaluation metrics are non-differentiable text-based measures~\cite{ranzato2015sequence}. Self-Critical Sequence Training (SCST)~\cite{Rennie_2017_CVPR} was proposed to address this issue by directly optimizing the inference output using caption-level rewards, such as CIDEr-D~\cite{Vedantam_2015_CVPR}.

There are two existing novel object captioning benchmarks: \emph{a)} the \emph{held-out} \coco Benchmark~\cite{Hendricks_2016_CVPR}, constructed by excluding images containing one of eight selected object classes from the standard \coco 2014 benchmark, and \emph{b)} \nocaps~\cite{nocaps2019}, which uses the \coco 2017 benchmark for training and provides new validation and test images from the \openimages Dataset with over 400 novel objects. Both benchmarks are object-centric and there is no reliable benchmarks that systematically evaluate the quality of generated actions or attributes.

\section{Model}
\begin{figure*}[!ht]
\centering
\includegraphics[width=\textwidth]{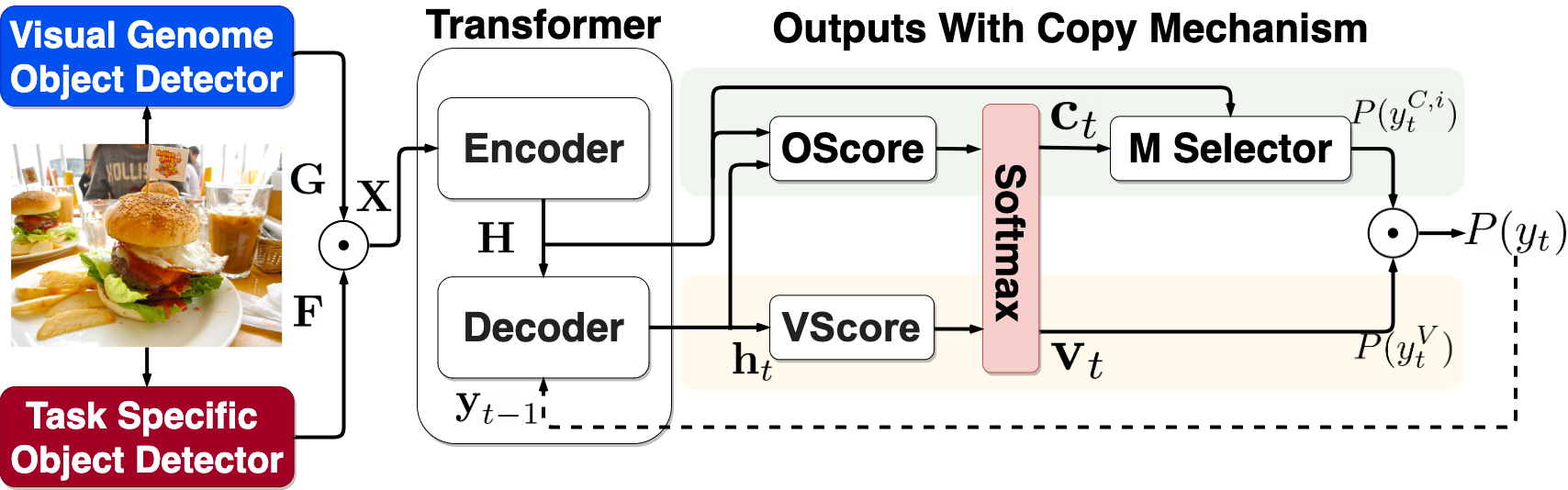}
\caption{Overview of the \nbtpp Model. $X$ is the concatenated Object representations $G$ and $F$ from the two object detectors. The Transformer encoder produces $\mathbf{H}$ and the decoder provides $h_t$ at step $t$. We then estimate the probabilities for generating each vocabulary word (yellow box) and copying from each \copyable image object (green box). The results are concatenated and jointly softmax (red box). We refine each copy probability into the concrete inflected word probability in $\mathit{MSelector}$. The final output $P(y_t)$ concatenates all above word probabilities.} 
\label{overview}
\end{figure*}

Figure~\ref{overview} provides an overview of the \nbtpp model. We refer to the \nbtpp model without SCST training as \nbtppnor. We describe this model in Sec.~\ref{objinput} and our novel reinforced copy encouragement training in Sec.~\ref{scst}.

\subsection{The \nbtppnor Model}
\label{objinput}

\paragraph{Input Image Objects:}
Following the setup in~\newcite{nocaps2019}, we use two object detectors: the \noncopyable object detector from~\newcite{Anderson_2018_CVPR}, producing image objects and regions $G$ (represented by embedding vectors $[\vx^g_1, \dots, \vx^g_{k^g}]$) with detailed visual features; and a \copyable object detector, producing image objects $F$ (represented by $[\vx^f_{1}, \dots, \vx^f_{k^f}]$) and their corresponding labels $L_f = [l_1, \dots, l_{k^{f}}]$ used as copy candidates during caption generation. We will introduce object representations $\vx_i$ below and define them in Eq.~\ref{objrep}.

\paragraph{Image Object Representations:}
Following~\newcite{Anderson_2018_CVPR,lu2018neural}, we represent both sets of objects with Region-Of-Interest (ROI, $\vr_{i} \in {\R}^{2048}$) vectors from the \noncopyable object detector and object positional features ($\vp_{i} \in {\R}^{8}$), including bounding box coordinates and size, and an object label confidence score. In addition, to transfer knowledge from the seen objects to the novel ones, we propose \emph{Abstract Labels} for the \copyable objects, described below.

\begin{figure}[!ht]
\centering
    \includegraphics[width=0.48\textwidth]{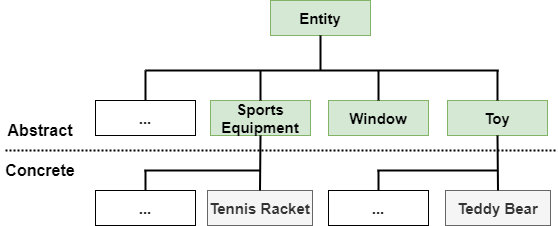}
\caption{A part of the class hierarchy from the Open Images V4 Dataset~\cite{kuznetsova2018open}. The green nodes are used as abstract object labels. For each label, its abstract label is its closest green ancestor.}
\label{objtree}
\end{figure}

\paragraph{Abstract Labels:}
The \copyable object detectors we use provide taxonomies of object classes, and every detected object is assigned a label from that taxonomy. More general object classes conceptually include all the labels lower in the taxonomy.~\footnote{If the \copyable object detector does not provide such a taxonomy, a suitable taxonomy could be obtained from sources such as Wordnet.} This provides us with a mechanism for associating class labels not present in the training data with those that do occur in the training data by mapping them to a common ancestor in the hierarchy. Inspired by~\newcite{ciaramita-johnson-2003-supersense}, we define \emph{Abstract Labels} to be a fixed set of ancestor class labels that spans the entire taxonomy (see Figure~\ref{objtree}). Using the abstract labels to drive copy decisions allows the usage of known object types to inform the word generation of novel objects. Each object from the \copyable detector is associated with its nearest abstract label ancestor. We choose the set of abstract labels such that the objects in the training data are evenly distributed across the set of abstract labels. We represent abstract labels with trainable embeddings $\ve_i \in \R^d$, where $d$ is the hidden size of our base model. We use the \openimages V4 class hierarchy for the \nocaps benchmark and a merged 8 coco super-categories hierarchy~\newcite{Chen2015} for the \emph{held-out} COCO benchmark. 
The final representation for each object $\vx_i$ is:
\begin{align}
\vx_i = \mathit{LN}(\mW_{r} \vr_{i} + \ve_{i}) + \mathit{LN}(\mW_{p} \vp_{i})
\label{objrep}
\end{align}
where $\mathit{LN}$ is layer normalization and $\mW_{r} \in \R^{d \times 2048}$, $\mW_{p} \in \R^{d \times 8}$ are trainable projections. The two sets of object representations are concatenated as $X = F \odot G$ where $\odot$ represents concatenation.

\paragraph{Transformer Base Model:}
We use a transformer model~\cite{NIPS2017_7181} with an $N_{enc}$-layer encoder and an $N_{dec}$-layer decoder ($N_{enc}$ = $N_{dec}$ = 3 in our experiments). We denote the encoder output $\mathbf{H} = \mathit{Encoder}(X)$. The decoder uses frozen word and positional embeddings $\mathit{WE}$ and $\mathit{PE}$ from GPT2~\cite{radford2019language} which are helpful in producing better captions describing novel objects. In step $t$:
\begin{align}
\vw_{1:t-1} &= \mathit{WE}(y_{1:t-1}) + \mathit{PE}(y_{1:t-1}) \\
\vh_t &= \mathit{Decoder}(\mathbf{H}, \vw_{1:t-1})
\end{align}
where $y_{1:t-1}$ is the generation history and $\vh_t \in \R^{d}$.

\paragraph{Outputs With Copy Mechanism:}
The \nbtppnor model either generates words from the vocabulary or copies from \copyable objects. We deploy a copy mechanism similar to the dynamic pointer network in~\newcite{Hu_2020_CVPR}. Given the decoder output $\vh_t$, we first calculate a raw score for each vocabulary word:
\begin{align}
\mathit{VScore}(\vh_t) &= \mW_e \vh_t
\label{vocabscore}
\end{align}
where $\mW_e \in \R^{|V| \times d}$, $|V|$ is the GPT2 vocabulary size. We then calculate raw additive attention scores over the encoder output of \copyable image objects (i.e., $\mathbf{H}_{1:k^f}$):
\begin{align}
\scalebox{0.95}{$\mathit{OScore}(\mathbf{H}, \vh_{t})_i = \vw^T_{c} \mathit{tanh}(\mW_{f}\mathbf{H}_i + \mW_{h}\vh_{t})$}
\end{align}
where $i \in [1, k^f]$ and $\mW_{f} \in R^{d \times d}$, $\mW_h \in R^{d \times d}$ and $\vw_{c} \in R^{d}$. Finally, we concatenate the raw scores from $\mathit{VScore}$ and $\mathit{OScore}$ and jointly softmax:
\begin{align}
\scalebox{0.83}{$[\vv_t, \vc_t] = \mathit{Softmax}([\mathit{VScore}(\vh_t) \odot \mathit{OScore}(\mathbf{H}, \vh_{t})])$}
\label{eq:combined-softmax}
\end{align}
where $\odot$ represents concatenation. $\vv_t$ provides probabilities for GPT2 vocabulary words and $\vc_t$ provides probabilities for copying \copyable object labels.

\paragraph{Morphological Selector:}
\label{morph-selector}
Object labels can appear in inflected forms in captions. For example, in Figure~\ref{example}, after selecting the object \emph{hamburger}, the \nbtppnor model should generate \emph{``hamburgers''} after \emph{``Two''}. We propose a morphological selector (\emph{M Selector}) to refine the copy probability of each \copyable image object label $l_i$ (i.e., $\vc_{t, i}$) into the probabilities of generating all possible morphological forms $y_t^{l_i}$ (i.e., $P(y_t^{l_i}|l_i)$). Specifically, we use $\vh_t$ to choose an inflected form from its possible inflected forms (e.g., Singular or Plural in English):
\begin{align}
P(y_t^{l_i}|l_i) &=  \softmax(\mW_{l_i} \vh_t) \label{eq:p-obj} 
\end{align}
Here $\mW_{l_i} \in \R^{s_i \times d}$ where $s_i$ is the number of inflected forms of label $l_i$ (in most cases $2$ for English, singular and plural). Finally, the \nbtppnor model concatenates the above refined probabilities as follows:
\begin{align}
P(y^v_t) &= \vv_t \\
P(y^{l_i}_{t}) &= \vc_{t, i} \cdot P(y_t^{l_i}|l_i) \label{eqn:P-y_l}\\
P(y_t) &= P(y^{v}_t) \odot P(y^{l_1}_{t}) \odot  \dots \odot P(y^{l_{k^f}}_{t})
\end{align}
where $\odot$ represents concatenation. Some novel object labels are included in the GPT2 vocabulary. However, these words are not present in the training captions and thus the model always assigns them very low probabilities in $P(y^v_t)$. The only way novel object labels can appear in captions is through copy operations.

\paragraph{Model Application Scope}
In this paper, we focus on the Novel Object Captioning task. However, in general, our copy mechanism is capable of copying any type of information. The Abstract Label approach is general to zero shot learning problems where novel items share characteristics with training instances. The Morphological Selector is also applicable to linguistic copy mechanisms in other contexts such as Commonsense Reasoning~\cite{lin-etal-2020-commongen} where copied terms may require linguistic alignment with the generated text.

\subsection{Copying More Object Labels}
\label{scst}
In this paper, we encourage the copying of object labels by using a suitable reward function in the Self-Critical Sequence Training (SCST) framework, which has proven effective for image captioning tasks. Compared with injecting additional loss terms together with the standard XE loss, using the SCST framework allows us to design arbitrary encouragement signals based on the inference output.  It minimizes the negative expected reward score:
\begin{align}
L_{R}(\theta) &= -\expectwrt[y_{1:T} \sim p_\theta]{r(y_{1:T})}
\end{align}
where $r$ is the reward function and $p_\theta$ represents the models outputs. In this paper, following~\newcite{cornia2020m2}, we first pre-train the \nbtppnor model with the CE loss, then switch to fine-tune the \nbtppnor model with the above SCST loss.

\paragraph{Inference Bias Baseline:}
We add an Inference Bias (IB) $b \in \R^+$ to increase $P(y^{l_i}_{t})$ at inference time. Eq.~\ref{eqn:P-y_l} is changed to:
\begin{align}
P(y^{l_i}_{t}) &= b \cdot\vc_{t, i} \cdot P(y_t^{l_i}|l_i)
\end{align}
and remaining probabilities normalised accordingly. IB is functionally equivalent to adjusting the threshold for the copy decision during inference. Surprisingly, this simple inference trick provides a strong baseline (see Table~\ref{ablation}). This shows that after the CE training, many correct copy operations are assigned with low probabilities, compared to the fixed vocabulary items. However, we believe that it is better to train the model to increase the probabilities of these copy operations than adding ad hoc adjustments at inference time. 

\paragraph{Can Standard SCST Encourage Copying?}
\newcite{Rennie_2017_CVPR} shows that SCST with the CIDEr reward fine-tuning leads to noticeable caption quality improvement for standard image captioning tasks (i.e, improvement in various automatic metrics). Previous work~\cite{cornia2020m2,Anderson_2018_CVPR} use CIDEr as the standard reward function in their SCST optimization. This shows suggests that the problem of overfitting of SCST training with CIDEr reward is minimal. Intuitively, the CIDEr reward is positively correlated with the number of salient object label mentions and should encourage the model to copy salient novel object labels. However, CIDEr equally rewards both generation of object labels present in training data via the vocabulary $P(y^v_t)$ and via copy operations $P(y^{l_i}_{t})$. Novel objects labels however can only be generated by copy operations (see Sec.~\ref{morph-selector}), thus the CIDEr reward function does not sufficiently encourage these operations. We propose two orthogonal modifications to the standard SCST algorithm to address this issue:

\paragraph{Novel Encouragement Reward:}
We propose combining the standard CIDEr-D reward with a reward function that gives captions with words copied from object labels an extra bonus, which we intend to encourage copy operations. One straightforward way to implement this idea is to provide a constant bonus to each triggered copy operation:
\begin{align}
\label{constant-bounds}
\var{R_a}(X) = \var{CIDEr}(X) + a * C
\end{align}
where $X$ is a generated caption, $C$ is the number of copy actions in the caption $X$ and $a \in R^{+}$ is a fixed hyper-parameter. We refer this as \emph{additive bias}. Optimizing with the additive bias, the captioning model only needs to trigger the copy operation for arbitrary objects at arbitrary generation steps. That is, the model may encourage copying object labels 
at the expense of caption quality (i.e., high CIDEr-D scores). Therefore, we propose a \emph{proportional bias} that assigns different rewards to the copy operations in different images by making a connection between the copy bonus and the generated captions CIDEr-D score:
\begin{align}
\label{connect-bounds}
\var{R_p}(X) = \var{CIDEr}(X) * (1.0 + p * C)
\end{align}
where $p \in R^{+}$ is a fixed hyper-parameter. Although $\var{R_a}$ can effectively encourage the model to copy objects, it may introduce noisy object mentions. $\var{R_p}$ penalizes those noisy object mentions via the low caption CIDEr score.

\paragraph{Visual Object Aligned (VOA) Images:}
\emph{VOA Images} refers to the set of training images where the reference captions contain at least one word from retained object labels. During SCST training, images that contain no object label words (i.e., non-VOA images) will not utilise copy operations, thus these images encourage the model NOT to copy. VOA images account for approximately 70\% of the full COCO-2017 training images set. Although restricting training to VOA images can be done with arbitrary models, this may hurt the diversity of generated captions. Experiments in Table~\ref{ablation} confirm that restricting to VOA images only improves performance when used with SCST training. 

\paragraph{Hyper-Parameters For Copy Encouragement:} 
The above approaches introduce two additional parameters: $a$ and $p$. In our experiments, $a$ and $p$ range over 0.2, 0.3 and 0.4; we found that 0.3 works the best for both reward functions. Combined with restricting SCST training to VOA images, $\var{R_p}$ works better than $\var{R_a}$ and sets a new SOTA for novel object image captioning.

\section{Experiments}
\begin{table*}
\begin{center}
\footnotesize
\setlength\tabcolsep{1pt}
\renewcommand{\arraystretch}{1.2}
\begin{tabularx}{\textwidth}{lYYYYYYYYYYYYYY} %
\toprule
& \multicolumn{2}{c}{\nocapsin} & \multicolumn{2}{c}{\nocapsnear} & \multicolumn{2}{c}{\nocapsout} & \multicolumn{3}{c}{\ttbf{Overall}} \\

Method  & CIDEr & SPICE & CIDEr & SPICE &  CIDEr & SPICE & Meteor & CIDEr & SPICE \\ 
\midrule
\updown + BS     & 73.7 & 11.6 &  57.2 & 10.3 & 30.4  & 8.1  & 22.9 & 54.5 & 10.1 \\
\updown + ELMo + CBS & 76.0 & 11.8 & 74.2 & 11.5 & 66.7 &  9.7 & 24.4 & 73.1 & 11.2 \\
\nbt + BS     & 62.8 & 10.3 & 51.9 & 9.4 & 48.9 & 8.4 & 21.8 & 54.3 & 9.4 \\
\nbt + CBS & 61.9 & 10.4 & 57.3 & 9.6 & 61.8 & 8.6 & 21.6 & 59.9 & 9.5 \\
$OSCAR_L$ + CBS + SCST & - & - & - & - & - & - & - & 80.9 & 11.3 \\
\bottomrule
\nbtppnor + IB (Ours) & 81.7 & 12.9 & 77.2 & 12.1 & 67.0 & 10.3 & 25.6 & 76.0 & 11.9 \\
\nbtpp  (Ours) & \textbf{87.3} & \textbf{12.8} & \textbf{84.0} & \textbf{12.5} & \textbf{75.4} & \textbf{10.7} & \textbf{25.7} & \textbf{82.9} & \textbf{12.2}  \\
\bottomrule
\end{tabularx}
\caption{Comparison of the \nbtpp model with other state-of-the-art systems on the \nocaps Test Split. The \nbtpp model sets new state of the art and improves previous work by 2.0 CIDEr and 0.9 SPICE. The performance of the \updown and \nbt models are from~\newcite{nocaps2019}. The $OSCAR_L$ model is from~\newcite{li2020oscar}.}
\label{mainresulttest}
\end{center}
\end{table*}

We conduct experiments on the \nocaps~\cite{nocaps2019} and the \emph{held-out} \coco~\cite{Hendricks_2016_CVPR} Benchmark. We set the layer and embedding size to $d$ = 768 and use Adam optimisation~\cite{kingma2014adam}. We train our models 15 epochs with batch size 100 for CE loss and 15 epochs with batch size 10 for SCST loss.

\begin{table*}[!ht]
\begin{center}
\footnotesize
\setlength\tabcolsep{1pt}
\renewcommand{\arraystretch}{1.2}
\begin{tabularx}{\textwidth}{cYYYYcYYY} %lccYYYY
\toprule
& \multicolumn{4}{c}{\nocapsout} & & \multicolumn{3}{c}{\nocapsin} \\
 \cmidrule{2-5} \cmidrule{7-9}
Method  & Meteor & CIDEr & SPICE & Object F1 && Meteor & CIDEr & SPICE \\ 
\midrule  
LSTM-P~\cite{Li_2019_CVPR} & 23.4 & 88.3 & 16.6 & 60.9 && - & - & -\\
Base + CBS~\cite{anderson-etal-2017-guided} & 23.3 &  77.0 & 15.9 & 54.0 && 24.9 & 88.0 & 18.4 \\
NBT + CBS~\cite{lu2018neural} & 24.1 & 86.0 & 17.4 & 70.5 &&   25.0      &     92.1 & 18.0  \\  
\midrule
\nbtppnor + IB (Ours) & 25.6 & 95.5   & 18.8 & 58.2 &&  \textbf{27.0} & 108.3 & 20.4 \\ 
\nbtpp (Ours) & \textbf{25.7} & \textbf{99.2} & \textbf{19.3} & 66.3 && 26.8 & \textbf{113.3} & 20.4  \\ 
\nbtpp + CBS (Ours) & \textbf{25.7} & 99.1 & 19.1 & \textbf{71.8} && 26.8 & 112.6 & \textbf{20.8}  \\ 
\midrule
PS3~\cite{anderson2018partially}$^{\S}$ & 25.4 & 94.5 & 17.9 & 63.0    &&  25.9  &  101.1 &  19.0 \\
FDM-net~\cite{caofeature}$^{\S}$ & 25.9 & 84.8 & 19.4 & 64.7 && 27.2 & 109.7 & 20.2 \\
FDM-net + CBS~\cite{caofeature}$^{\S}$ & 25.6 & 85.3 & 19.6 & 85.7 && 26.2 & 105.5 & 19.7 \\
\bottomrule
\end{tabularx}
\caption{Comparison of the \nbtpp model with previous work on the test split of the \emph{held-out} \coco Benchmark. Object F1 measures the presence of particular novel objects in the captions. $^{\S}$ not comparable as they use additional information related to novel objects (i.e., images containing novel objects and scene graphs).}
\label{eightcocoperformance}
\end{center}
\end{table*}

\subsection{Evaluation Metrics}
We use CIDEr~\cite{Vedantam_2015_CVPR}, SPICE~\cite{spice2016} and METEOR~\cite{banerjee2005meteor} to evaluate the caption quality. CIDEr measures the similarity between the reference captions and generated outputs using tf-idf weighted n-gram overlap. SPICE is based on the scene graphs matching between the reference captions and generated outputs. METEOR focuses on the alignment between the words in reference captions and generated outputs, with an aim of 1:1 correspondence. To measure the effectiveness of our copy encouragement approach, we report object F1~\cite{anderson-etal-2017-guided} in the \emph{held-out} \coco Benchmark. As the \nocaps benchmark does not release its ground-truth captions, we instead report averaged number of mentioned objects (Ave. O) and CIDEr score for dummy captions that only contain copied object words (Object CIDEr, OC., details see Appendix).

\subsection{Comparison with the State-of-the-art}
We compare our models \nbtppnor + IB and \nbtpp with other state-of-the-art systems in Tables~\ref{mainresulttest} and~\ref{eightcocoperformance}. 

On the \nocaps benchmark (Table~\ref{mainresulttest}), 
our models outperform previous work, including the recently proposed $OSCAR_L$ + CBS + SCST model~\cite{li2020oscar}, which is fine-turned from the BERT-LARGE model~\cite{devlin-etal-2019-bert}, by 2.0 CIDEr, 0.9 SPICE and set a new state of the art. Compared with the $OSCAR_L$ model, our models use far fewer model parameters (340M vs. 60M) and outperforms $OSCAR_L$ on both CIDEr and SPICE metrics. We train our model for about 10 hours for CE Loss and 24 hours for SCST Loss using a single Nvidia P100 GPU. As a comparison, the $OSCAR_L$ model which is fine-tuned from BERT-LARGE uses 60 - 90 hours for training CE Loss and 60 - 90 hours for training SCST Loss.~\footnote{According to the authors' comments on their official model code repo \url{https://github.com/microsoft/Oscar/issues/6}} This shows that simply deploying a BERT-based language model is not sufficient for the Novel Object Captioning task. 

On the \emph{held-out} \coco benchmark (Table~\ref{eightcocoperformance}), the \nbtpp model produces more novel objects (+ 13.3 Object F1) and  higher quality captions  (+ 3.9 CIDEr on the \nocapsout split) than the \nbtppnor model with run-time Inference Bias. Compared with previous work, the \nbtpp model achieves 10.9 CIDEr and 1.9 SPICE higher in the \nocapsout split, 21.2 CIDEr and 2.8 SPICE higher in the \nocapsin split with the highest object F1. This shows that our copy encouragement approach successfully trains our model to correctly copy more novel objects and to produce high-quality captions. Compared with PS3~\cite{anderson2018partially} and FDM-net model~\cite{caofeature} which are trained on extra images containing novel objects and scene graphs, our models still outperform the PS3 model and 13.9 CIDEr higher than the FDM-net. We set a new state of the art in this benchmark without additional novel objects information.

\begin{table*}[!ht]
\centering
\footnotesize
\setlength\tabcolsep{2pt}
\renewcommand{\arraystretch}{1.2}
\begin{tabularx}{\textwidth}{lYYYYYYYYYY}
\toprule
& \multicolumn{2}{c}{\nocapsin} & \multicolumn{2}{c}{\nocapsnear} & \multicolumn{2}{c}{\nocapsout} & \multicolumn{4}{c}{\ttbf{Overall}} \\
Method  &  CIDEr &  SPICE &  CIDEr &  SPICE &  CIDEr &  SPICE &  CIDEr &  SPICE & OC. & Ave. O\\ 
\midrule
\specialrule{.4pt}{0pt}{0pt}
\emph{\footnotesize{Ablation of \nbtppnor Components}} \\ [-0.6ex]
\textbf{(1)} R + P + AL & 80.3 & 12.7 & 60.3 & 11.4 & 34.8 & 9.2 & 58.0 & 11.2 & - & - \\
\textbf{(2)} GPT2 + R + P + AL & 78.6 & 12.5 & 64.2 & 11.6 & 39.4 & 9.3 & 61.3 & 11.3 & - & - \\
\textbf{(3)} C + GPT2 + R + P + AL (\nbtppnor) & 80.7 & 12.7 & 68.6 & 11.8 & 61.0 & 10.2 & 68.8 & 11.7 & 10.3 & 0.5 \\
% \textbf{(4)} \nbtppnor w/o AL & 79.6 & 12.3 & 67.9 & 11.4 & 60.4 & 9.7 & 68.1 & 11.2 & 11.5 & 0.6 \\
% \textbf{(5)} \nbtppnor w/o SP & xx & xx & xx & xx & 57.5 & 10.2 & 67.0 & 11.1 & 10.0 & 0.5 \\
\specialrule{.4pt}{0pt}{0pt}
\emph{\footnotesize{Ablation of Copy Encouragement}} \\ [-0.6ex]
\textbf{(4)} \nbtppnor + IB & 85.3 & 12.8 & 76.6 & 12.2 & 73.7 & 10.8 & 77.2 & 12.0 & 17.5 & 1.1 \\
\textbf{(5)} \nbtppnor + CIDEr & 89.7 & 12.3 & 75.9 & 11.8 & 63.5 & 10.1 & 75.3 & 11.5 & 11.7 & 0.6 \\
\textbf{(6)} \nbtppnor + $\var{R_a}$  & 89.0 & 12.6 & 82.2 & 12.5 & 78.2 & 11.2 & 82.3 & 12.3 & 19.7 & 1.5 \\
\textbf{(7)} \nbtppnor+ $\var{R_p}$  & 92.3 & 12.9 & 83.5 & 12.5 & 74.9 & 11.0 & 83.0 & 12.3 & 17.7 & 1.3 \\
\textbf{(8)} \nbtppnor + $\var{R_a}$ w/ VOA & 89.3 & 12.6 & 83.6 & 12.5 & 81.7 & 11.1 & 84.0 & 12.2 & 20.7 & 1.6 \\
\textbf{(9)} \nbtppnor + $\var{R_p}$ w/ VOA (all training) & 88.0 & 12.6 & 83.3 & 12.6 & 79.4 & 11.2 & 83.2 & 12.3 & 20.6 & 1.5 \\
\textbf{(10)} \nbtppnor + $\var{R_p}$ w/ VOA (\nbtpp) & 90.4 & 12.6 & 84.5 & 12.4 & 83.1 & 11.0 & 85.1 & 12.2 & 20.4 & 1.5 \\
\bottomrule
\end{tabularx}
\caption{Ablation study of our model. \emph{OC.} for Object CIDEr; \emph{Ave. O} for Averaged Number of Mentioned Object in each image; \emph{C} for Copy Mechanism; \emph{R} for ROI; \emph{P} for Position; \emph{AL} for Abstract Label; $\var{R_a}$ and  $\var{R_p}$ are the SCST reward function; \emph{VOA}: Visual Object Aligned Images; \emph{VOA (all training)}: all training using VOA images.}
\label{ablation}
\end{table*}

\subsection{Ablation Study}
Table~\ref{ablation} presents ablation results for various \nbtpp components, including our copy encouragement approach. Table~\ref{performanceIB} shows that our encouragement of copying in the \nbtpp model does not benefit from additional Inference Bias. 
Table~\ref{alsp} shows the effect of Abstract Labels and the Morphological Selector in the \nbtpp model. Finally, Table~\ref{cocoperformance} confirms the \nbtpp model's generalization ability for in-domain \coco images.

\paragraph{\nbtpp Components:}
The \nbtppnor model produces better captions using the frozen GPT2 parameters (row \textbf{1} vs. \textbf{2}). Our copy mechanism (C) helps the model to explicitly integrate novel objects, substantially improving the \nocapsout split by 15.3 CIDEr and 0.3 SPICE (row \textbf{2} vs. \textbf{3}). The Inference Bias (IB) introduces noticeable performance improvement: 8.4 CIDEr and 0.3 SPICE (row \textbf{3} vs. \textbf{4}) in models that do not use our reinforcement learning approach. The \nbtppnor model trained with the standard SCST reward function obtains an overall 8.1 CIDEr improvement, but most of the improvement is from the \nocapsin and \nocapsnear splits (row \textbf{8} vs. \textbf{6}). Compared with the \nbtppnor + IB model, the \nbtppnor model trained with standard SCST algorithm is 8.1 CIDEr lower in the \nocapsout split (row \textbf{5} vs. \textbf{4}). As discussed in Sec.~\ref{scst}, standard SCST cannot provide sufficient copy encouragement as object words can be generated from either pathways (fixed vocabulary or copy). Optimizing either the $\var{R_a}$ or $\var{R_p}$ reward functions improves the \nbtppnor + CIDEr model by 7.0 CIDEr and 7.8 CIDEr respectively (row \textbf{7} and \textbf{5}). $\var{R_a}$ achieves 3.7 CIDEr higher than $\var{R_p}$ in the \nocapsout split. Interestingly, after restricting the model training to VOA images, $\var{R_p}$ achieves 7.8 CIDEr improvement in the \nocapsout split (row \textbf{8} vs. \textbf{7}), outperforming the \nbtppnor + $\var{R_a}$ w/ VOA model by 1.4 CIDEr (row \textbf{10} vs. \textbf{8}). 

\begin{figure*}[t!]
	\centering
	\footnotesize
	\setlength{\tabcolsep}{0.6em}
	\begin{tabularx}{\textwidth}{p{2cm}XXX}
	    & \multicolumn{1}{c}{\textbf{\emph{curtain}}} & \multicolumn{1}{c}{\textbf{\emph{ostrich, deer}}} & \multicolumn{1}{c}{\textbf{\emph{door, house}}} \\
        & \includegraphics[width=0.27\textwidth,height=2.7cm]{image/1013.jpg}&
		\includegraphics[width=0.27\textwidth,height=2.7cm]{image/1714.jpg} & 
		\includegraphics[width=0.27\textwidth,height=2.7cm]{image/1723.jpg}\\

		\cmidrule{1-4}
		\nbtpp & A bathroom with a shower \underline{\textbf{\textit{curtain}}} and a toilet.\checkmark &
    	\underline{\textbf{\textit{An ostrich}}} and a deer standing in a field. \checkmark&
		A red door of a red house with a red phone. $\times$ \\
		\nbtppnor + IB & A white bath tub sitting next to a white toilet. $\times$ &
	    Two ostriches and a deer in a grassy field. $\times$ &
		A red telephone booth sitting next to a brick wall. \checkmark\\
		\cmidrule{1-4}
		GT & The bathtub is white and has a white shower \underline{\textbf{\textit{curtain}}}. &
    	\underline{\textbf{\textit{An ostrich}}} standing in grass with a few deer in the background. &
    	A red phone booth is standing against a brick wall. \\

	\end{tabularx}
    \caption{Three Examples generated by the \nbtpp and \nbtppnor + IB model on the \nocaps Val Set. First Case: The \nbtpp model accurately mentions the novel object \emph{curtain} in the caption. Second Case: Both models talk about \emph{ostrich} in their generated caption, but the \nbtppnor + IB model uses the wrong modifier. Third Case: when detected object labels are too general, the \nbtpp model may produce inaccurate captions.}

	\label{fig:nocaps_model_predictions}
\end{figure*}

\paragraph{Effectiveness of Copy Encouragement:}
We directly measure the copy quantity by counting the number of copied object labels and Object CIDEr. Row \textbf{5} and \textbf{3} confirm that the standard SCST algorithm has little impact on the copy quantity (only + 0.1 object per image and + 1.4 Object CIDEr). Inference Bias (IB), $\var{R_a}$ and $\var{R_p}$ rewards substantially improve the quantity of copied objects (row \textbf{4}, \textbf{6}, \textbf{7} vs. \textbf{3}). Among these three components, the models trained with $\var{R_a}$ and $\var{R_p}$ work better than the IB baseline (row \textbf{6}, \textbf{7} vs. \textbf{4}). The model trained with the $\var{R_a}$ reward copies more objects than the $\var{R_p}$ reward, especially training with all training images. This is because the $\var{R_a}$ reward assigns constant positive reward for all copied objects. However, such a naive reward appears to encourage noisy copying operations (i.e., copying non-salient objects). As a result, the \nbtppnor + $\var{R_a}$ model performs worse than the \nbtppnor + $\var{R_p}$ model in terms of caption quality (row \textbf{7} vs. \textbf{6}). After restricting training with VOA images, the models trained with $\var{R_a}$ and $\var{R_p}$ copy similar amount of objects, but the model with $\var{R_a}$ produce better captions than the one with $\var{R_a}$, especially in the \nocapsout split (row \textbf{10} vs. \textbf{8}). The $\var{R_p}$ reward maintains a good balance between copying more objects and high caption quality.

\paragraph{Are The VOA images Always Useful?}
Restricting training to the VOA images can be done with any captioning models. However, this does not necessarily encourage copy operations and improve the output caption quality. When we restrict training to VOA images, the \nbtpp model performs consistently worse in all three splits compared to our proposed training scheme (row \textbf{9} vs. \textbf{10}). The only difference is that the \nbtppnor model is not trained with diverse images during the cross-entropy stage. That is, restricting to VOA images is only suitable for fine-tuning in the SCST stage.

\paragraph{Sufficient Encouragement For Copy:}
Here we investigate whether our \nbtpp model mentions a sufficient number of salient objects. We apply an increasing amount of inference bias to the \nbtppnor, \nbtppnor + CIDEr and \nbtpp models in Table~\ref{performanceIB}. We note that only \nbtpp model is negatively impacted (measured by CIDEr score) by different Inference Bias values. This shows that the \nbtpp model does not benefit from further copy encouragement.

\begin{table}[!ht]
\begin{center}
\footnotesize
\setlength\tabcolsep{1pt}
\renewcommand{\arraystretch}{1.2}
\begin{tabularx}{0.5\textwidth}{cYYYY} %lccYYYY
\toprule
IB  & 1 ($e^0$) & 2.72 ($e^1$) & 7.39 ($e^2$) & 20.01 ($e^3$) \\ 
\midrule
\nbtppnor & 68.8 & 73.6 & 76.0  & 77.2  \\
\nbtppnor + CIDEr & 75.3 & 77.6 & 79.5 & 81.1 \\
\nbtpp & \textbf{85.1} & 85.0 & 84.9   & 84.3 \\
\bottomrule
\end{tabularx}
\caption{The CIDEr Score in \nocaps Validation Set with different Inference Bias.}
\label{performanceIB}
\end{center}
\end{table}

\paragraph{The Effect of Abstract Labels and M Selector}
Table~\ref{alsp} shows the effect of Abstract Labels (AL) and the M Selector (M) in the \nbtppnor + IB and \nbtpp models. Removing AL and M from the \nbtppnor + IB model drops 2.3 CIDEr, 0.6 SPICE and 4.6 CIDEr, 0.8 SPICE respectively. AL and M have a large positive impact on the SPICE score. As SPICE is sensitive to long-range object word relationships, such as attributes and predicate words,~\cite{spice2016} Abstract Labels and the M Selector improve the semantic coherence and fluency of the captions. The performance gap in the \nbtpp model becomes smaller. Our copy encouragement approach contributes to the generation coherency and fluency. 

\begin{table}[!ht]
\centering
\footnotesize
\setlength\tabcolsep{2pt}
\renewcommand{\arraystretch}{1.2}
\begin{tabularx}{0.48\textwidth}{lYYYY}
\toprule
 & \multicolumn{2}{c}{\nocapsout} & \multicolumn{2}{c}{\ttbf{Overall}} \\
Method  &  CIDEr &  SPICE &  CIDEr &  SPICE \\ 
\midrule
% \nbtppnor & 61.0 & 10.2 & 68.8 & 11.7 \\
% \nbtppnor w/o AL & 60.4 & 9.7 & 68.1 & 11.2 \\
% \nbtppnor w/o SP & 57.5 & 10.2 & 67.0 & 11.1 \\
% \specialrule{.4pt}{0pt}{0pt}
\nbtppnor + IB & 73.7 & 10.8 & 77.2 & 12.0 \\
\nbtppnor + IB w/o AL & 72.7 & 10.0 & 74.9 & 11.4 \\
\nbtppnor + IB w/o M & 71.2 & 10.5 & 72.6 & 11.2 \\
\specialrule{.4pt}{0pt}{0pt}
\nbtpp & 83.1 & 11.0 & 85.1 & 12.2 \\
\nbtpp w/o AL & 80.8 & 10.9 & 84.4 & 12.1 \\
\nbtpp w/o M & 78.4 & 10.7 & 82.8 & 11.9 \\
\bottomrule
\end{tabularx}
\caption{The contribution of Abstract Label (AL) and Morphological Selector (M).}
\label{alsp}
\end{table}

\begin{table}[!ht]
\begin{center}
\footnotesize
\setlength\tabcolsep{1pt}
\renewcommand{\arraystretch}{1.2}
\begin{tabularx}{0.5\textwidth}{cYYY} %lccYYYY
\toprule

Method  & Meteor & CIDEr & SPICE \\ 
\midrule
\nbt + CBS & 25.1 & 80.2 & 15.8 \\
\updown + CBS & 25.7 & 95.4 & 18.2 \\
\midrule
\nbtppnor + IB & 27.1 & 106.1   & 20.6  \\ 
\nbtpp & 27.1 & 114.2   & 20.9 \\ 

\bottomrule
\end{tabularx}
\caption{The performance of the best four \nocaps models on the \coco 2017 Validation Set.}
\label{cocoperformance}
\end{center}
\end{table}

\paragraph{Generalization For In-Domain \coco:}
To further show the generalization of our model for In-Domain images (i.e., without novel objects), we run the \nbtppnor + IB and \nbtpp models on the \coco 2017 Validation Set and compare with another two novel object captioning models (\nbt + CBS and \updown + ELMo + CBS) reported in~\newcite{nocaps2019} in Table~\ref{cocoperformance}. Both of our models outperforms the \updown and \nbt model by a large margin. Our models produce high-quality captions for images with novel objects as well as known objects.

\subsection{Qualitative analysis on \nocaps}
Qualitative analysis on the \nocaps validation set reveals that the \nbtpp model mentions the salient object in the input image (first example in Figure~\ref{fig:nocaps_model_predictions}), is able to generate more accurate descriptions of novel objects (second example in Figure~\ref{fig:nocaps_model_predictions}), however may generate inaccurate captions due to the non-informative detected object labels (third example in Figure~\ref{fig:nocaps_model_predictions}). In summary, the \nbtpp model is better at incorporating detected image objects into generated captions than the \nbtppnor + IB model.

\section{Conclusion and Future work}

This paper proposes the \nbtpp model that includes a training scheme to encourage copying novel object labels using Reinforced Learning. Our experiments show that the \nbtpp model successfully integrates novel object information and achieves state-of-the-art performance on two Novel Object Caption Benchmarks. In the future, we plan to extend our SCST reward function to other metrics such as SPICE~\cite{anderson2016spice} and BertScore~\cite{Zhang2020BERTScore}.  

\section*{Acknowledgments}
We thank anonymous reviewers for their insightful suggestions to improve this paper. This research was supported by a Google award through the Natural Language Understanding Focused Program, by a MQ Research Excellence Scholarship and a CSIRO’s DATA61 Top-up Scholarship, and under the Australian Research Councils Discovery Projects funding scheme (project number DP160102156). 
% The acknowledgments should go immediately before the references. Do not number the acknowledgments section.
% Do not include this section when submitting your paper for review.

\bibliography{anthology,emnlp2020}

\begin{thebibliography}{33}
\expandafter\ifx\csname natexlab\endcsname\relax\def\natexlab#1{#1}\fi

\bibitem[{Agrawal et~al.(2019)Agrawal, Desai, Wang, Chen, Jain, Johnson, Batra,
  Parikh, Lee, and Anderson}]{nocaps2019}
Harsh Agrawal, Karan Desai, Yufei Wang, Xinlei Chen, Rishabh Jain, Mark
  Johnson, Dhruv Batra, Devi Parikh, Stefan Lee, and Peter Anderson. 2019.
\newblock nocaps: novel object captioning at scale.
\newblock In \emph{Proceedings of the IEEE/CVF International Conference on
  Computer Vision (ICCV)}.

\bibitem[{Anderson et~al.(2016{\natexlab{a}})Anderson, Fernando, Johnson, and
  Gould}]{spice2016}
Peter Anderson, Basura Fernando, Mark Johnson, and Stephen Gould.
  2016{\natexlab{a}}.
\newblock Spice: Semantic propositional image caption evaluation.
\newblock In \emph{ECCV}.

\bibitem[{Anderson et~al.(2016{\natexlab{b}})Anderson, Fernando, Johnson, and
  Gould}]{anderson2016spice}
Peter Anderson, Basura Fernando, Mark Johnson, and Stephen Gould.
  2016{\natexlab{b}}.
\newblock Spice: Semantic propositional image caption evaluation.
\newblock In \emph{European Conference on Computer Vision}, pages 382--398.
  Springer.

\bibitem[{Anderson et~al.(2017)Anderson, Fernando, Johnson, and
  Gould}]{anderson-etal-2017-guided}
Peter Anderson, Basura Fernando, Mark Johnson, and Stephen Gould. 2017.
\newblock \href {https://doi.org/10.18653/v1/D17-1098} {Guided open vocabulary
  image captioning with constrained beam search}.
\newblock In \emph{Proceedings of the 2017 Conference on Empirical Methods in
  Natural Language Processing}, pages 936--945, Copenhagen, Denmark.
  Association for Computational Linguistics.

\bibitem[{Anderson et~al.(2018{\natexlab{a}})Anderson, Gould, and
  Johnson}]{anderson2018partially}
Peter Anderson, Stephen Gould, and Mark Johnson. 2018{\natexlab{a}}.
\newblock Partially-supervised image captioning.
\newblock In \emph{Advances in Neural Information Processing Systems}, pages
  1875--1886.

\bibitem[{Anderson et~al.(2018{\natexlab{b}})Anderson, He, Buehler, Teney,
  Johnson, Gould, and Zhang}]{Anderson_2018_CVPR}
Peter Anderson, Xiaodong He, Chris Buehler, Damien Teney, Mark Johnson, Stephen
  Gould, and Lei Zhang. 2018{\natexlab{b}}.
\newblock Bottom-up and top-down attention for image captioning and visual
  question answering.
\newblock In \emph{The IEEE Conference on Computer Vision and Pattern
  Recognition (CVPR)}.

\bibitem[{Banerjee and Lavie(2005)}]{banerjee2005meteor}
Satanjeev Banerjee and Alon Lavie. 2005.
\newblock Meteor: An automatic metric for mt evaluation with improved
  correlation with human judgments.
\newblock In \emph{Proceedings of the acl workshop on intrinsic and extrinsic
  evaluation measures for machine translation and/or summarization}, pages
  65--72.

\bibitem[{Cao et~al.(2020)Cao, Han, Wang, Ma, Fu, Jiang, and Xue}]{caofeature}
Tingjia Cao, Ke~Han, Xiaomei Wang, Lin Ma, Yanwei Fu, Yu-Gang Jiang, and
  Xiangyang Xue. 2020.
\newblock Feature deformation meta-networks in image captioning of novel
  objects.
\newblock In \emph{Proceedings of the AAAI Conference on Artificial
  Intelligence}.

\bibitem[{Ciaramita and Johnson(2003)}]{ciaramita-johnson-2003-supersense}
Massimiliano Ciaramita and Mark Johnson. 2003.
\newblock \href {https://www.aclweb.org/anthology/W03-1022} {Supersense tagging
  of unknown nouns in {W}ord{N}et}.
\newblock In \emph{Proceedings of the 2003 Conference on Empirical Methods in
  Natural Language Processing}, pages 168--175.

\bibitem[{Cornia et~al.(2020)Cornia, Stefanini, Baraldi, and
  Cucchiara}]{cornia2020m2}
Marcella Cornia, Matteo Stefanini, Lorenzo Baraldi, and Rita Cucchiara. 2020.
\newblock {Meshed-Memory Transformer for Image Captioning}.
\newblock In \emph{Proceedings of the IEEE/CVF Conference on Computer Vision
  and Pattern Recognition}.

\bibitem[{Devlin et~al.(2019)Devlin, Chang, Lee, and
  Toutanova}]{devlin-etal-2019-bert}
Jacob Devlin, Ming-Wei Chang, Kenton Lee, and Kristina Toutanova. 2019.
\newblock \href {https://doi.org/10.18653/v1/N19-1423} {{BERT}: Pre-training of
  deep bidirectional transformers for language understanding}.
\newblock In \emph{Proceedings of the 2019 Conference of the North {A}merican
  Chapter of the Association for Computational Linguistics: Human Language
  Technologies, Volume 1 (Long and Short Papers)}, pages 4171--4186,
  Minneapolis, Minnesota. Association for Computational Linguistics.

\bibitem[{Hendricks et~al.(2016)Hendricks, Venugopalan, Rohrbach, Mooney,
  Saenko, and Darrell}]{Hendricks_2016_CVPR}
Lisa~Anne Hendricks, Subhashini Venugopalan, Marcus Rohrbach, Raymond~J.
  Mooney, Kate Saenko, and Trevor Darrell. 2016.
\newblock \href {https://doi.org/10.1109/CVPR.2016.8} {Deep compositional
  captioning: Describing novel object categories without paired training data}.
\newblock In \emph{2016 {IEEE} Conference on Computer Vision and Pattern
  Recognition, {CVPR} 2016, Las Vegas, NV, USA, June 27-30, 2016}, pages 1--10.

\bibitem[{Herdade et~al.(2019)Herdade, Kappeler, Boakye, and
  Soares}]{NIPS2019_9293}
Simao Herdade, Armin Kappeler, Kofi Boakye, and Joao Soares. 2019.
\newblock \href
  {http://papers.nips.cc/paper/9293-image-captioning-transforming-objects-into-words.pdf}
  {Image captioning: Transforming objects into words}.
\newblock In H.~Wallach, H.~Larochelle, A.~Beygelzimer, F.~dAlch\'{e} Buc,
  E.~Fox, and R.~Garnett, editors, \emph{Advances in Neural Information
  Processing Systems 32}, pages 11137--11147. Curran Associates, Inc.

\bibitem[{Hu et~al.(2020)Hu, Singh, Darrell, and Rohrbach}]{Hu_2020_CVPR}
Ronghang Hu, Amanpreet Singh, Trevor Darrell, and Marcus Rohrbach. 2020.
\newblock Iterative answer prediction with pointer-augmented multimodal
  transformers for textvqa.
\newblock In \emph{Proceedings of the IEEE/CVF Conference on Computer Vision
  and Pattern Recognition (CVPR)}.

\bibitem[{Kingma and Ba(2014)}]{kingma2014adam}
Diederik~P Kingma and Jimmy Ba. 2014.
\newblock Adam: A method for stochastic optimization.
\newblock \emph{arXiv preprint arXiv:1412.6980}.

\bibitem[{Kuznetsova et~al.(2018)Kuznetsova, Rom, Alldrin, Uijlings, Krasin,
  Pont-Tuset, Kamali, Popov, Malloci, Duerig et~al.}]{kuznetsova2018open}
Alina Kuznetsova, Hassan Rom, Neil Alldrin, Jasper Uijlings, Ivan Krasin, Jordi
  Pont-Tuset, Shahab Kamali, Stefan Popov, Matteo Malloci, Tom Duerig, et~al.
  2018.
\newblock The open images dataset v4: Unified image classification, object
  detection, and visual relationship detection at scale.
\newblock \emph{arXiv preprint arXiv:1811.00982}.

\bibitem[{Li et~al.(2020)Li, Yin, Li, Hu, Zhang, Zhang, Wang, Hu, Dong, Wei,
  Choi, and Gao}]{li2020oscar}
Xiujun Li, Xi~Yin, Chunyuan Li, Xiaowei Hu, Pengchuan Zhang, Lei Zhang, Lijuan
  Wang, Houdong Hu, Li~Dong, Furu Wei, Yejin Choi, and Jianfeng Gao. 2020.
\newblock Oscar: Object-semantics aligned pre-training for vision-language
  tasks.
\newblock \emph{ECCV}.

\bibitem[{Li et~al.(2019)Li, Yao, Pan, Chao, and Mei}]{Li_2019_CVPR}
Yehao Li, Ting Yao, Yingwei Pan, Hongyang Chao, and Tao Mei. 2019.
\newblock Pointing novel objects in image captioning.
\newblock In \emph{The IEEE Conference on Computer Vision and Pattern
  Recognition (CVPR)}.

\bibitem[{Lin et~al.(2020)Lin, Zhou, Shen, Zhou, Bhagavatula, Choi, and
  Ren}]{lin-etal-2020-commongen}
Bill~Yuchen Lin, Wangchunshu Zhou, Ming Shen, Pei Zhou, Chandra Bhagavatula,
  Yejin Choi, and Xiang Ren. 2020.
\newblock \href {https://www.aclweb.org/anthology/2020.findings-emnlp.165}
  {{C}ommon{G}en: A constrained text generation challenge for generative
  commonsense reasoning}.
\newblock In \emph{Findings of the Association for Computational Linguistics:
  EMNLP 2020}, pages 1823--1840, Online. Association for Computational
  Linguistics.

\bibitem[{Lin et~al.(2014)Lin, Maire, Belongie, Hays, Perona, Ramanan,
  Doll{\'a}r, and Zitnick}]{Chen2015}
Tsung-Yi Lin, Michael Maire, Serge Belongie, James Hays, Pietro Perona, Deva
  Ramanan, Piotr Doll{\'a}r, and C.~Lawrence Zitnick. 2014.
\newblock Microsoft coco: Common objects in context.
\newblock In \emph{Computer Vision -- ECCV 2014}, pages 740--755, Cham.
  Springer International Publishing.

\bibitem[{Lu et~al.(2018{\natexlab{a}})Lu, Yang, Batra, and
  Parikh}]{lu2018neural}
Jiasen Lu, Jianwei Yang, Dhruv Batra, and Devi Parikh. 2018{\natexlab{a}}.
\newblock Neural baby talk.
\newblock In \emph{Proceedings of the IEEE Conference on Computer Vision and
  Pattern Recognition}, pages 7219--7228.

\bibitem[{Lu et~al.(2018{\natexlab{b}})Lu, Ni, Ji, Sakamoto, Shibuki, and
  Mori}]{lu-etal-2018-deep}
Yujie Lu, Boyi Ni, Qijin Ji, Kotaro Sakamoto, Hideyuki Shibuki, and Tatsunori
  Mori. 2018{\natexlab{b}}.
\newblock \href {https://www.aclweb.org/anthology/Y18-1049} {Deep learning
  paradigm with transformed monolingual word embeddings for multilingual
  sentiment analysis}.
\newblock In \emph{Proceedings of the 32nd Pacific Asia Conference on Language,
  Information and Computation}, Hong Kong. Association for Computational
  Linguistics.

\bibitem[{Radford et~al.(2019)Radford, Wu, Child, Luan, Amodei, and
  Sutskever}]{radford2019language}
Alec Radford, Jeffrey Wu, Rewon Child, David Luan, Dario Amodei, and Ilya
  Sutskever. 2019.
\newblock Language models are unsupervised multitask learners.
\newblock \emph{OpenAI Blog}, 1(8):9.

\bibitem[{Ranzato et~al.(2015)Ranzato, Chopra, Auli, and
  Zaremba}]{ranzato2015sequence}
Marc'Aurelio Ranzato, Sumit Chopra, Michael Auli, and Wojciech Zaremba. 2015.
\newblock Sequence level training with recurrent neural networks.
\newblock \emph{arXiv preprint arXiv:1511.06732}.

\bibitem[{Rennie et~al.(2017)Rennie, Marcheret, Mroueh, Ross, and
  Goel}]{Rennie_2017_CVPR}
Steven~J. Rennie, Etienne Marcheret, Youssef Mroueh, Jerret Ross, and Vaibhava
  Goel. 2017.
\newblock Self-critical sequence training for image captioning.
\newblock In \emph{The IEEE Conference on Computer Vision and Pattern
  Recognition (CVPR)}.

\bibitem[{Vaswani et~al.(2017)Vaswani, Shazeer, Parmar, Uszkoreit, Jones,
  Gomez, Kaiser, and Polosukhin}]{NIPS2017_7181}
Ashish Vaswani, Noam Shazeer, Niki Parmar, Jakob Uszkoreit, Llion Jones,
  Aidan~N Gomez, \L~ukasz Kaiser, and Illia Polosukhin. 2017.
\newblock \href
  {http://papers.nips.cc/paper/7181-attention-is-all-you-need.pdf} {Attention
  is all you need}.
\newblock In I.~Guyon, U.~V. Luxburg, S.~Bengio, H.~Wallach, R.~Fergus,
  S.~Vishwanathan, and R.~Garnett, editors, \emph{Advances in Neural
  Information Processing Systems 30}, pages 5998--6008. Curran Associates, Inc.

\bibitem[{Vedantam et~al.(2015)Vedantam, Lawrence~Zitnick, and
  Parikh}]{Vedantam_2015_CVPR}
Ramakrishna Vedantam, C.~Lawrence~Zitnick, and Devi Parikh. 2015.
\newblock Cider: Consensus-based image description evaluation.
\newblock In \emph{The IEEE Conference on Computer Vision and Pattern
  Recognition (CVPR)}.

\bibitem[{Vinyals et~al.(2015)Vinyals, Fortunato, and Jaitly}]{NIPS2015_5866}
Oriol Vinyals, Meire Fortunato, and Navdeep Jaitly. 2015.
\newblock \href {http://papers.nips.cc/paper/5866-pointer-networks.pdf}
  {Pointer networks}.
\newblock In C.~Cortes, N.~D. Lawrence, D.~D. Lee, M.~Sugiyama, and R.~Garnett,
  editors, \emph{Advances in Neural Information Processing Systems 28}, pages
  2692--2700. Curran Associates, Inc.

\bibitem[{Wang et~al.(2019)Wang, Chen, and Hu}]{wang2019hierarchical}
Weixuan Wang, Zhihong Chen, and Haifeng Hu. 2019.
\newblock Hierarchical attention network for image captioning.
\newblock In \emph{Proceedings of the AAAI Conference on Artificial
  Intelligence}, volume~33, pages 8957--8964.

\bibitem[{Wu et~al.(2018)Wu, Zhu, Jiang, and Yang}]{wu2018decoupled}
Yu~Wu, Linchao Zhu, Lu~Jiang, and Yi~Yang. 2018.
\newblock \href {https://doi.org/10.1145/3240508.3240640} {Decoupled novel
  object captioner}.
\newblock In \emph{2018 {ACM} Multimedia Conference on Multimedia Conference,
  {MM} 2018, Seoul, Republic of Korea, October 22-26, 2018}, pages 1029--1037.

\bibitem[{Yadav et~al.(2019)Yadav, Jain, Agrawal, Chattopadhyay, Singh, Jain,
  Singh, Lee, and Batra}]{yadav2019evalai}
Deshraj Yadav, Rishabh Jain, Harsh Agrawal, Prithvijit Chattopadhyay, Taranjeet
  Singh, Akash Jain, Shiv~Baran Singh, Stefan Lee, and Dhruv Batra. 2019.
\newblock Evalai: Towards better evaluation systems for ai agents.
\newblock \emph{arXiv preprint arXiv:1902.03570}.

\bibitem[{Yao et~al.(2017)Yao, Pan, Li, and Mei}]{yao2017incorporating}
Ting Yao, Yingwei Pan, Yehao Li, and Tao Mei. 2017.
\newblock Incorporating copying mechanism in image captioning for learning
  novel objects.
\newblock In \emph{Proceedings of the IEEE Conference on Computer Vision and
  Pattern Recognition}, pages 6580--6588.

\bibitem[{Zhang et~al.(2020)Zhang, Kishore, Wu, Weinberger, and
  Artzi}]{Zhang2020BERTScore}
Tianyi Zhang, Varsha Kishore, Felix Wu, Kilian~Q. Weinberger, and Yoav Artzi.
  2020.
\newblock \href {https://openreview.net/forum?id=SkeHuCVFDr} {Bertscore:
  Evaluating text generation with bert}.
\newblock In \emph{International Conference on Learning Representations}.

\end{thebibliography}
\bibliographystyle{acl_natbib}

\newpage
\begin{appendices}

\newcommand{\dmodel}{d_{\text{model}}}

\section{Model Details}

The hyper-parameters of the \nbtpp model is shown in Table~\ref{modelhp}. This architecture is basically from~\cite{cornia2020m2}. We only change the hidden size of the model to 768 to fit the size of GPT2 (the smallest version). Our model has total $60.8 \times 10^6$ parameters and $43.0 \times 10^6$ trainable parameters. This scale is slightly smaller than the Transformer Base model ($65.8 \times 10^6$)~\cite{NIPS2017_7181}. 

We optimise with Adam($\alpha$=0.9, $\beta$=0.98, $\epsilon$=1e-9)~\cite{kingma2014adam} and clip gradients to 0.1 for both Benchmarks. In Cross-entropy training, we vary the learning rate over the course of training using the heuristic:
\begin{equation}
lrate = \dmodel^{-0.5} \cdot \min({S}^{-0.5}, S \cdot W^{-1.5})
\end{equation}
where $S$ is the step number and $W$ is the number of warm-up steps. We set $W$ to 20000 steps for the \nocaps Benchmark and 10000 steps for the \emph{held-out} COCO Benchmark. The number of warm-up steps has some impact on both benchmark. We tried 20,000 and 10,000 for both Benchmarks.  For SCST training, we set the initial learning rate $1e^{-6}$ and reduce it by half if the reward metric (Validatoin set CIDEr) does not improve for 3 evaluations. We conduct evaluation every 3000 steps. 

We use Pytorch \emph{1.4.0} to implement our model. The Cross-Entropy Training takes about 8 hours and the SCST optimization takes about 15 hours in a single NVIDIA Tesla P100 GPU. 

Our source code is submitted in the \emph{Software}. We setup an anonymous Google Drive to host large file~\footnote{\url{https://drive.google.com/drive/folders/1EToBXQ8WAWxn5uCd38HtfRYmchnBBbMo?usp=sharing}}.
\begin{table}[!ht]
\begin{tabular}{|c|c|c|c|}
\hline
Name        & Value & Name               & Value \\ \hline
$N_e$       & 3     & Dropout            & 0.1   \\ \hline
$N_d$       & 3     & Max Cap. Len. & 20    \\ \hline
Model Size & 768   & beam Size     & 5   \\ \hline
FFN Size    & 2048  & Att. Head Num & 8     \\ \hline
\end{tabular}
\caption{Hyper-parameters of the \nbtpp model.}
\label{modelhp}
\end{table}

\subsection{Input Object Detector}
We follow the processing of input objects in~\newcite{nocaps2019}. We observed that some object categories are frequently mentioned in the training captions and that they often have variable, context-sensitive verbalisation (e.g., a person might be described as a sports player, a student, etc., depending on the context). For those objects, vocabulary based word generation often did a better job at selecting the correct verbalisation due to their frequency in training captions. On the other hand, novel objects typically have lower-frequencies and a fixed, single verbalisation. For example, elephants are usually only referred to with the word \emph{elephant}. For this reason, we remove objects with high-frequency in training captions from the output of the \copyable object detector, leaving their corresponding words to be generated via vocabulary softmax. We also remove the more abstract objects (higher in the object class hierarchy) when object regions overlap. Finally, we keep only one detected object for each label (the one with highest confidence score). 

We provide the downloadable link of filtered results in Sec~\ref{datadetail}. We use exactly the same \noncopyable objects as described in~\newcite{Anderson_2018_CVPR}. The \noncopyable object detector~\cite{Anderson_2018_CVPR} can produce ROI vectors for arbitrary bounding boxes, hence we use it also to produce ROI vectors for objects from the \copyable detector.

\subsection{\nbtpp Inference Details}
We use Beam Search with beam size 5 to decode the captions. We first do length normalization for the overall score of each decoded caption. We also penalize captions when they generate repeated bi-grams. Once the repetitions are found, the log-probability for that particular word is divided by a penalty value $e^2$. All image objects are only allowed to be copied once. During the SCST optimization, we mask out words from the vocabulary that can be generated via copy operations to encourage the model to copy. All the above constraints are applied to all of our models in the ablation study.

\subsection{Object CIDEr Details}
Object CIDEr score for dummy captions that only contain copied object words. This shows the correctness of our copy mechanism. High Object CIDEr score means many of the copied object labels are also mentioned in the ground-truth captions. We use this metrics because the \nocaps benchmark does not release its ground-truth captions and only provide online evaluation APIs.  

\begin{table*}[t!]
\begin{center}
\footnotesize
\setlength\tabcolsep{1pt}
\renewcommand{\arraystretch}{1.2}
\begin{tabularx}{\textwidth}{cYYYYcYYY} %lccYYYY
\toprule
& \multicolumn{4}{c}{\nocapsout} & & \multicolumn{3}{c}{\nocapsin} \\
 \cmidrule{2-5} \cmidrule{7-9}
Method  & Meteor & CIDEr & SPICE & Object F1 && Meteor & CIDEr & SPICE \\ 
\midrule  
\nbtppnor + IB (Ours) & 25.1 & 93.0   & 18.7 & 58.4 &&  26.9 & 107.1 & 20.3 \\ 
\nbtpp (Ours) & 25.6 & 99.2 & 19.5 & 64.5 && 26.7 & 113.0 & 20.3  \\ 
\nbtpp + CBS (Ours) & 25.6 & 98.4 & 19.3 & 70.9 && 26.7 & 112.3 & 20.2  \\ 
\bottomrule
\end{tabularx}
\caption{The performance of \nbtppnor + IB, \nbtpp and \nbtpp + CBS model on \emph{held-out} COCO Benchmark Validation Set.}
\label{eightcocoperformanceval}
\end{center}
\end{table*}

\section{Dataset Details}
\label{datadetail}
For the \nocaps Benchmark, we train with the COCO-2017 dataset, which is available at \url{http://images.cocodataset.org/zips/train2017.zip}. The \nocaps Validation and Test datasets are available from \url{https://nocaps.org/download}.
The \noncopyable image object detection files can be found in \url{https://github.com/nocaps-org/updown-baseline}. For the \emph{held-out} COCO Benchmark, the training and evaluation data can be found in \url{https://github.com/LisaAnne/DCC}. The \noncopyable image object detector is used for both benchmarks because COCO-2017 and COCO-2014 share the same set of images. The anonymous Google Drive includes the above data and the sets of \copyable objects detected for the above two benchmarks.

\subsection{Duplicated Caption Removal}
We find some images in COCO share exactly the same reference captions. We find it beneficial to remove those duplicates. We simply iterate over all reference captions and remove any captions if they have already been found previously. This removes 25463 captions from the training data of the \nocaps Benchmark and 7059 captions from the training data of the \emph{held-out} COCO Benchmark. 

\subsection{VOA (visual object aligned) Images}
VOA (visual object aligned) images/reference caption pairs are those that mention at least one detected \copyable image object label (or their linguistic variant). Non-VOA image/caption pairs are removed from our SCST training process. We provide the reduced set of reference captions in the anonymous Google Drive (\emph{ddc\_captions/ddc\_train\_VOA.json} and \emph{nocaps\_captions/nocaps\_train\_VOA.json}).

\begin{table}[!ht]
\begin{center}
\footnotesize
\setlength\tabcolsep{1pt}
\renewcommand{\arraystretch}{1.2}
\begin{tabularx}{0.5\textwidth}{cYYYYY} %lccYYYY
\toprule
         & COCO Train & COCO Train VOA & COCO Val & \nocaps Val & \nocaps Test \\
\midrule
\#Image   & 118,287          & 82,771               & 5,000          & 4,500       & 10,600       \\
\#Caption & 591,753          & 299,502              & 25,014         & -          & -          \\
\bottomrule
\end{tabularx}
\caption{Data Statistics for the \nocaps Benchmark. The full set of annotations in \nocaps Val and Test is not available. One can only access some of them via \url{https://nocaps.org/explore}.}
\label{nocapsstatics}
\end{center}
\end{table}

\subsection{Data Statistics for Benchmarks}
Table~\ref{nocapsstatics} and Table~\ref{ddcstatics} show the number of images and annotated reference captions of the \nocaps and \emph{held-out} COCO Benchmark, respectively. On average, each image has five reference captions. The COCO Train in the \nocaps Benchmark is larger than the \emph{held-out} COCO Benchmark.

\begin{table}[!ht]
\begin{center}
\footnotesize
\setlength\tabcolsep{1pt}
\renewcommand{\arraystretch}{1.2}
\begin{tabularx}{0.5\textwidth}{cYYYYYY} %lccYYYY
\toprule
          & Train  & Train VOA & Val in-domain & Val out-domain & Test in-domain & Test out-domain \\
\midrule
\#Image   & 70,194  & 55,799     & 17,234         & 3,018           & 17,288          & 3,024     \\
\#Caption & 351,134 & 197,061    & 86,230         & 1,5105          & 86,188          & 15,131    \\
\bottomrule
\end{tabularx}
\caption{Data Statistics for the \emph{held-out} COCO Benchmark.}
\label{ddcstatics}
\end{center}
\end{table}

\section{Evaluation}
The \nocaps Benchmark hosts its evaluation sever at~\url{https://evalai.cloudcv.org/web/challenges/challenge-page/355/overview}. The detailed setup instruction of the local submission environment to Evai~\cite{yadav2019evalai} is available at \url{https://github.com/nocaps-org/updown-baseline}. The \emph{held-out} COCO Benchmark uses the evaluation tool from~\url{https://github.com/ruotianluo/coco-caption/tree/ea20010419a955fed9882f9dcc53f2dc1ac65092}. We provide an on-the-shelf version of this tool in the anonymous Google Drive (in \emph{tools}).

\subsection{\emph{held-out} COCO Benchmark Validation Performance}
We only show the test performance on the \emph{held-out} COCO Benchmark in our main paper. Here, we show the performance of our model performance on the validation Set in Table~\ref{eightcocoperformanceval}. The models achieve similar level of performance on the Validation Set.

\end{appendices}

\end{document}